\documentclass[a4paper,fleqn]{cas-dc}

\usepackage[numbers]{natbib}
\def\tsc#1{\csdef{#1}{\textsc{\lowercase{#1}}\xspace}}
\tsc{WGM}
\tsc{QE}
\usepackage{xcolor}
\usepackage[linesnumbered,lined,boxed,ruled]{algorithm2e}

\SetCommentSty{mycommfont}
\SetKwInput{KwInput}{Input}                % Set the Input
\SetKwInput{KwOutput}{Output}              % set the Output

\makeatletter
\newcommand{\removelatexerror}{\let\@latex@error\@gobble}
\makeatother

\begin{document}
\let\WriteBookmarks\relax
\def\floatpagepagefraction{1}
\def\textpagefraction{.001}

\shorttitle{A parameter-free graph reduction for spectral clustering and SpectralNet}    

\shortauthors{Alshammari et~al.}  

\title [mode = title]{A parameter-free graph reduction for spectral clustering and SpectralNet}  

%% Title footnote mark
%% eg: \tnotemark[1]
%\tnotemark[<tnote number>] 
%
%% Title footnote 1.
%% eg: \tnotetext[1]{Title footnote text}
%\tnotetext[<tnote number>]{<tnote text>} 

\author[1]{Mashaan Alshammari}[
orcid=0000-0002-7864-0011]
\cormark[1]
\ead{mashaan.awad1930@alum.kfupm.edu.sa}

\address[1]{Independent Researcher, Riyadh, Saudi Arabia}

\author[2]{John Stavrakakis}[]
\ead{john.stavrakakis@sydney.edu.au}

\author[2]{Masahiro Takatsuka}[]
\ead{masa.takatsuka@sydney.edu.au}

\address[2]{School of Computer Science, The University of Sydney, NSW 2006, Australia}

\cortext[1]{Corresponding author}

%% Footnote text


\begin{abstract}
Graph-based clustering methods like spectral clustering and SpectralNet are very efficient in detecting clusters of non-convex shapes. Unlike the popular $k$-means, graph-based clustering methods do not assume that each cluster has a single mean. However, these methods need a graph where vertices in the same cluster are connected by edges of large weights. To achieve this goal, many studies have proposed graph reduction methods with parameters. Unfortunately, these parameters have to be tuned for every dataset. We introduce a graph reduction method that does not require any parameters. First, the distances from every point $p$ to its neighbors are filtered using an adaptive threshold to only keep neighbors with similar surrounding density. Second, the similarities with close neighbors are computed and only high similarities are kept. The edges that survive these two filtering steps form the constructed graph that was passed to spectral clustering and SpectralNet. The experiments showed that our method provides a stable alternative, where other methods’ performance fluctuated according to the setting of their parameters.
\end{abstract}

% Keywords

\maketitle

% Main text
\section{Introduction}\label{Introduction}
%Unsupervised classification, which is also known as clustering, refers to the problem of partitioning $N$ points $\{1,2,...,N\}$ into $C$ disjoint subsets $\{1,2,...,C\}$ without a prior knowledge of their categorization \cite{RN306}. In supervised and semi supervised classification, the method predicts the category of a sample based on known categories around it and statistical measures such as bayesian rule. This is not the case in unsupervised classification where the method predicts the category of a sample only based on statistical measures. Most well-known clustering methods including $k$-means \cite{RN309} and generative mixture models \cite{RN315} would only detect clusters of convex geometric shapes \cite{RN410}. Thus, they fail when this assumption is violated. That is because their statistical measures are looking for the cluster mean with samples scattered around it.

The problem of detecting clusters of non-convex geometric shape, has been long studied in the literature of pattern recognition. The solutions of this problem could be broadly classified into two categories: kernel- and graph-based methods. Kernel-based methods attempt to map the points into a space where they can be separated. The embedding function $\phi: \mathbb{R}^D \rightarrow \mathbb{R}^M$ maps points from the original space to an embedding space. Defining the embedding function $\phi$ is usually unknown and could be computationally expensive \cite{RN411}. On the other hand, graph-based methods use the graph $G(V,E)$ whose set of vertices represents the data points and its set of edges represents the similarity between each pair of vertices. Finding non-convex clusters in a graph could be done in three ways: 1) by iteratively coarsening and partitioning the graph \cite{RN412}, 2) by performing spectral clustering \cite{RN421}, and 3) by feeding the graph $G(V,E)$ to a neural network (SpectralNet) \cite{RN360}. The first way of detecting clusters in a graph is iterative which involves two deficiencies: the risk of being trapped in a local minima and the need for a stopping condition. This makes spectral clustering and SpectralNet more appealing for studies conducting graph-based clustering.

%\begin{figure*}[!t]
%	\centering
%	\includegraphics[width=0.75\textwidth,height=20cm,keepaspectratio]{figs-01/Fig-01.pdf}	
%	\caption{Different methods’ attempts to reduce the graph size. (Best viewed in color)}
%	\label{Fig:Fig-01}
%\end{figure*}

Spectral clustering starts by constructing a graph $G(V,E)$. The sets of vertices $V$ and edges $E$ represent data points and their pairwise similarities. Spectral clustering detects clusters by performing eigen-decomposition on the graph Laplacian matrix $L$ and running $k$-means on its top eigenvectors \cite{RN233}. The computational bottleneck represented by the eigen-decomposition would cost the algorithm computations in the order of $\mathcal{O}(N^3)$ \cite{RN232}. This stimulated the research on reducing these computations by reducing the graph vertices and/or edges. However, the need for a memory efficient graph creates another problem related to the number of parameters associated with the process of graph construction. Deciding the number of reduced vertices and how the edges are eliminated would create several parameters that need careful tuning.

SpectralNet \cite{RN360} uses Siamese nets to learn affinities between data points. Then it feeds these affinities to a neural network to find a map function $F_\theta$, which maps the graph vertices $V$ to an embedding space where they can be separated using $k$-means. The Siamese nets expect the user to label which pairs are positive (similar) and which are negative (dissimilar). An unsupervised pairing uses the $k$-nearest neighbors, where the nearest neighbors are the positive pairs and the farthest neighbors are the negative pairs. The parameter $k$ requires manual tuning. It also restricts the number of edges to be exactly $k$, regardless of the surrounding density around the data point.

In her spectral clustering paper, von Luxburg \cite{RN234} wrote about the advantages of mutual k-nearest neighbor graph, and how it \textit{``tends not to connect areas of different density''}. She highlighted the need for having a \textit{``heuristic to choose the parameter k''}. We introduce a graph reduction method that does not require any parameters to produce a mutual graph with a reduced number of edges compared to the size of the full graph $E = N \times N$, where $N$ is number of all vertices. It initially finds the mean distance that best describes the density around a point. Then, it computes the pairwise similarities based on: 1) the distance between a pair of points, and 2) the mean distance of the surrounding density. Finally, we construct a mutual graph where a pair of vertices must be in each other’s nearest neighbors sets. We used two graph applications for the experiments: spectral clustering and SpectralNet. The proposed method provides a stable alternative compared to other methods where their performance was determined by the selected parameters.

Our main contribution in this work, is eliminating manually tuning parameters that affect the clustering accuracy when changed. The graph partitioning methods used in this work are spectral clustering \cite{RN234} and SpectralNet \cite{RN360}.

%The rest of this paper is organized as follows. Section \ref{RelatedWork} provides an overview of methods used to detect non-convex clusters. Section \ref{ProposedApproach} introduces the proposed method and its steps to reduce edges in the graph. In section \ref{Experiments} we describe the setup of the experiments and discuss the findings.
\section{Related work}
\label{RelatedWork}
The problem of detecting non-convex clusters has led to the development of numerous clustering methods. These methods have abandoned the assumption that a cluster has a single mean. Instead, they rely on pairwise similarities to detect clusters. Graph-based clustering involves two steps: 1) reducing the graph, and 2) partitioning the graph. The proposed method in this paper falls under graph construction methods.

Spectral clustering uses eigen-decomposition to map the points into an embedding space, then groups similar points. One of the important application of spectral clustering is subspace clustering \cite{Liu2013Robust,Elhamifar2013Sparse}. The performance of spectral clustering is determined by the similarity metric used to construct the affinity matrix $A$. The earlier works of subspace clustering used affinities based on principal angles \cite{Wolf2003Learning}. But recent studies have used sparse representation of points to measure the similarity \cite{Liu2013Robust,Peng2015Subspace,Peng2021Kernel}. Spectral clustering requires computations in order of $\mathcal{O}(N^3)$, due to the eigen-decomposition step. A straightforward solution to this problem is to reduce the size of the affinity matrix $A$. This can be done in two ways: 1) reducing the set of vertices $V$, or 2) reducing the set of edges $E$.
 
Reducing the number of vertices is done by placing representatives on top of the data points, and then using those representatives as graph vertices. Placing a representative could be done by sampling (like $k$-means++ \cite{RN240}) or by vector quantization (like self-organizing maps \cite{RN42}). A well-known method in this field is ``k-means-based approximate spectral clustering (KASP)''  proposed by Yan et al. \cite{RN232}. KASP uses $k$-means to place representatives. Other efforts by Tasdemir \cite{RN275} and Tasdemir et al. \cite{RN276} involved placing representatives using vector quantization, and a nice feature of these methods is that the pairwise similarities are computed during the vector quantization. The problem with these methods is the parameter $m$, which is the number of representatives. Specifically, how should we set $m$? And how would different values of $m$ affect the clustering results?

Reducing the graph edges could be done by setting the neighborhood conditions. For instance, let $p$ be the center of a ball $B(p,r)$ with radius $r$ and $q$ be the center of a ball $B(q,r)$. $p$ and $q$ are connected if and only if the intersection of $B(p,r)$ and $B(q,r)$ does not contain other points \cite{Marchette2004Random}. Such graphs are called Relative Neighborhood Graphs (RNGs). Correa and Lindstrom \cite{RN254} used a $\beta$-skeleton graph for spectral clustering. However, the parameter $\beta$ needs tuning. Alshammari et al. \cite{RN421} introduced a method to filter edges from a $k$-nearest neighbor graph. However, it still needs an influential parameter, which was the mean of the baseline distribution of distances $\mu_0$. Another local method to reduce the number of edges was proposed by Satuluri et al. \cite{Satuluri2011Local}. The authors measured the similarity between two vertices using adjacency lists overlap, a metric known in the literature as shared nearest neighbor similarity \cite{Jarvis1973Clustering}. A graph sparsification method based on effective resistance was proposed by Spielman and Srivastava \cite{Spielman2011Graph,Spielman2011Spectral}. Their method was theoretically solid, but the definition of effective resistance breaks the cluster structure of the graph. Vertices with more short paths have low effective resistance and the method disconnects them. Retaining the cluster structure of the graph requires connecting such vertices \cite{Satuluri2011Local}.

In spectral clustering, the obtained spectral embedding cannot be extended to unseen data, a task commonly known as out-of-sample-extension (OOSE). Several studies have proposed solutions to this problem. Bengio et al. \cite{Bengio2003Extensions} used Nystr\"{o}m method to approximate the eigenfunction for the new samples. But they have to check the similarity between the training and new samples \cite{Nie2011Spectral}. Alzate and Suykens \cite{Alzate2010Multiway} proposed binarizing the rows of eigenvectors matrix where each row corresponds to a single training data point. By counting row occurrences, one can find the $k$ most occurring rows, where each row represents an encode vector for a cluster. To label a test sample, its projection was binarized and it is assigned to the closest cluster based on the minimum Hamming distance between its projection and encoding vectors. Levin et al. \cite{Levin2018extension} proposed a linear least squares OOSE, which was very close to Bengio et al. \cite{Bengio2003Extensions} approach. They also proposed a maximum-likelihood OOSE that produces a binary vector $\overrightarrow{a}$ indicating whether the unseen sample has an edge to the training samples or not.

All previous methods that provided an out-of-sample-extension (OOSE) to spectral clustering have relied on eigen-decomposition, which becomes infeasible for large datasets. The newly proposed SpectralNet \cite{RN360} is different from spectral clustering in a way that it does not use eigen-decomposition step. Instead, SpectralNet passes the affinity matrix $A$ into a deep neural network to group points with high similarities. Yet, SpectralNet still needs a graph construction method. Previous SpectralNet works have used $k$-nearest neighbour graph, but they have to set the parameter $k$ manually. It also restricts the number of edges to be exactly $k$, regardless of the surrounding density around the data point. Dense clusters require more edges to be strongly connected. Strong connections ensure closer positions in the embedding space. Also, SpectralNet methods randomly choose the negative pairs. This random selection makes the method inconsistent in independent executions.

Considering the literature on reduced graphs for spectral clustering and SpectralNet, it is evident that they have two deficiencies. First, certain parameters are required to drive the graph reduction process. Second, the involvement of random steps makes these methods inconsistent over independent executions.
%\begin{figure}[!b]%[!htp]
%	\centering
%	\includegraphics[width=0.95\textwidth,height=20cm,keepaspectratio]{figs-01/Fig-02.pdf}	
%	\caption{Steps of the proposed method. (Best viewed in color)}
%	\label{Fig:Fig-02}
%\end{figure}

\section{Reducing the graph size without the need for parameters}
\label{ProposedApproach}

The motivation behind our work was to avoid the use of any parameters during the graph reduction. The input for our method is a $k$-nearest neighbor graph. Although this $k$-nn graph is sparsified, it still connects clusters with different densities. The value of $k$ has limited influence on the final graph because it is not final, and most of the unnecessary edges created by $k$-nn will be removed in the reduction process. The method starts by finding the value of $\sigma_p$ that best describes the local statistics around a randomly chosen point $p$. Then, it filters the edges with low weights. Finally, it checks the mutual agreement for each edge. %The steps in the proposed method are illustrated in Figure\ \ref{Fig:Fig-02}.

\begin{figure*}%[h]%[!htp]
	\centering
	\includegraphics[width=\textwidth,height=20cm,keepaspectratio]{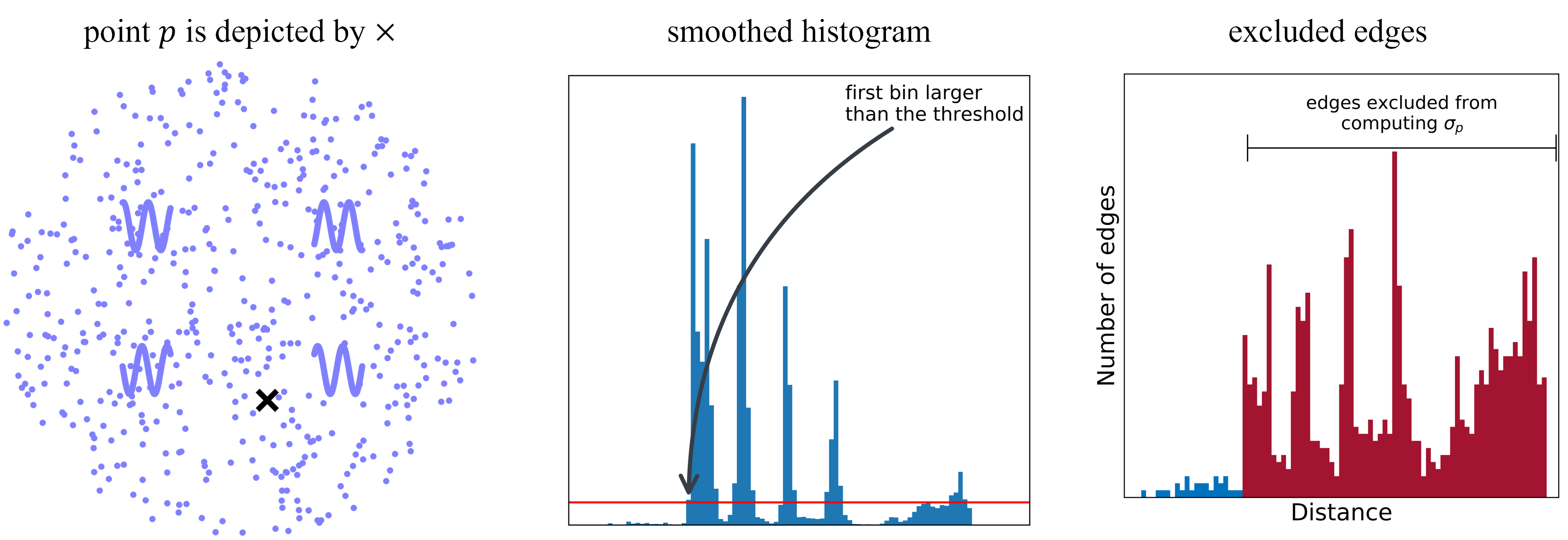}
	\caption{The process of computing $\sigma_p$ for a point $p$. (Best viewed in color)}
	\label{Fig:Fig-03}
\end{figure*}

\subsection{Finding the value of $\sigma_p$}
\label{sigmap}

To compute pairwise similarities, we used the similarity measure introduced by \cite{RN237}, which is defined as follows:
\begin{equation}
A_{pq}=\exp{\left(\frac{-d^2\left(p,q\right)}{\sigma_p \sigma_q\ }\right)}.
\label{Eq-003}
\end{equation}
\noindent
where $-d^2\left(p,q\right)$ is the distance between points $p$ and $q$. $\sigma_p$ and $\sigma_q$ are the local scales at points $p$ and $q$ respectively. What is good about this similarity measure is that it uses two sources of information to compute the pairwise similarity: 1) the distance between them, and 2) the surrounding density for each point. Points belonging to clusters with different densities would have a low similarity even if they are separated by a small distance This makes this measure superior for highlighting different clusters separated by a small distance.

One problem that arises from using this measure in equation \ref{Eq-003} is how to set the value of $\sigma_p$ in the denominator. In previous studies, it was set as the distance to the 7th neighbor \cite{RN237,RN274}. However, there is no evidence that the distance to the 7th neighbor would work in every dataset. Using the data to select this parameter would be more practical.

The idea behind the parameter $\sigma_p$ is to measure the sparseness of a cluster. If $p$ lies in a sparse cluster, it would have a large $\sigma_p$; whereas if $p$ lies in a dense cluster, it would have a small $\sigma_p$. To achieve this, we need to exclude neighbors with different local density than $p$ from being included in computing $\sigma_p$. We used a smooth histogram of distances to characterize the local density for $p$ neighbors (as shown in Figure\ \ref{Fig:Fig-03}). The intuition is if a neighbor has different local density than $p$, this would be represented as a peak on the histogram. The histogram bin values for each point are smoothed using the moving weighted average (MWA). The smoothing was designed as follows:
\begin{equation}
	MWA_{i}={\frac{v_{i-1}+v_{i}+v_{i+1}}{r_{i-1}+r_{i}+r_{i+1}}},
	\label{Eq-004}
\end{equation}
\noindent
where $v$ is the value of the bin and $r$ is the rank of the bin, with $r=1$ being the bin containing the closest points to the point $p$. This smoothing assigns weights to the bins based on their distance from $p$, with high weights assigned to closer bins and low weights assigned to bins further away.

The histogram threshold tells us that up to the Kth neighbor, the local density of $p$ has not changed. Then, we compute $\sigma_p$ as the mean distance from the 1st to the Kth neighbor. This process is described in statements 4 to 9 in Algorithm \ref{Alg:Alg-01}.

\subsection{Reducing the graph edges}
\label{ReduceGraphEdge}

\begin{figure*}%[!t]%[!htp]
	\centering
	\includegraphics[width=0.7\textwidth,height=20cm,keepaspectratio]{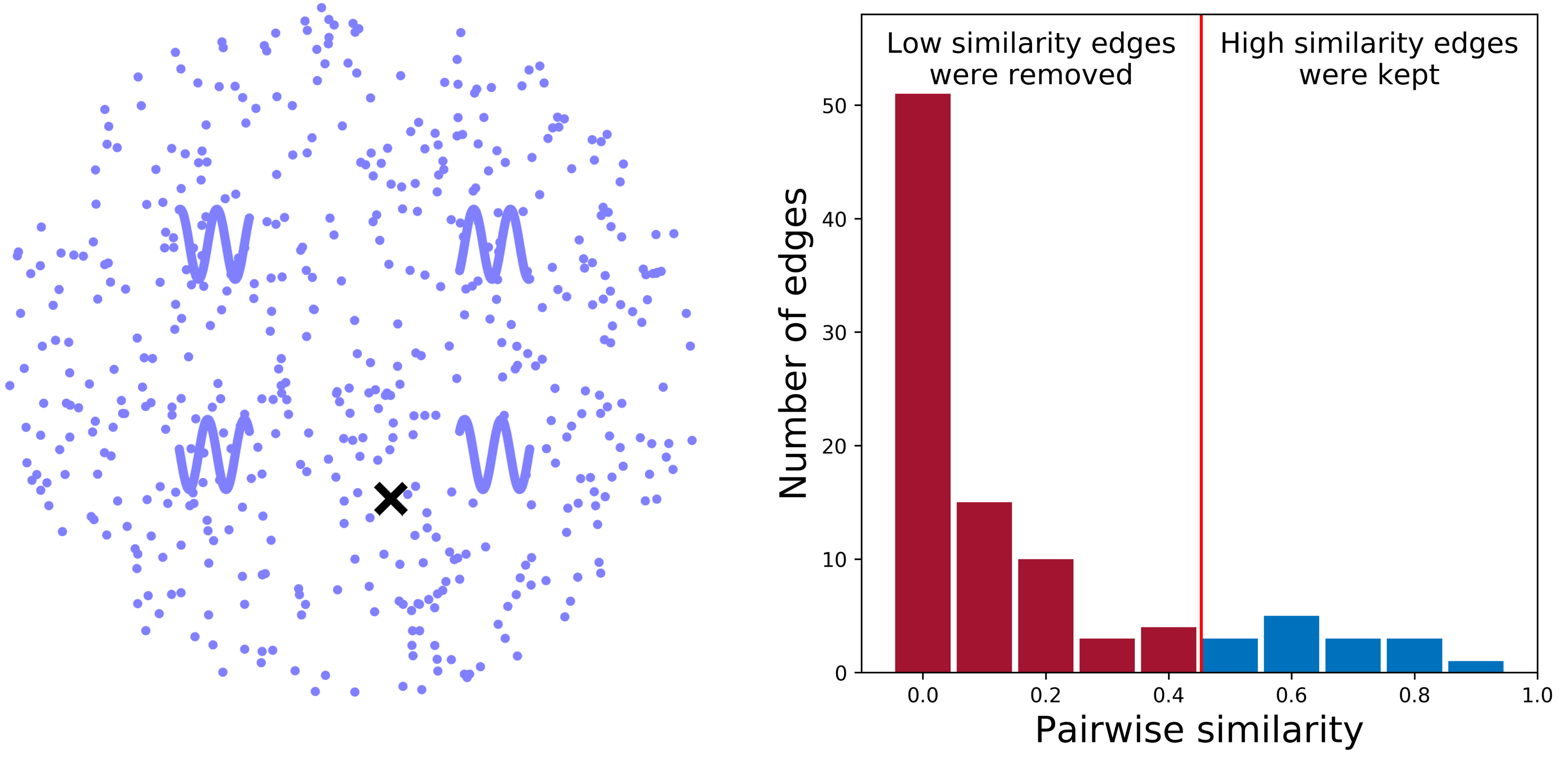}	
	\caption{After computing the pairwise similarities, we include highly similar edges for a point $p$. (Best viewed in color)}
	\label{Fig:Fig-05}
\end{figure*}

Once we have $\sigma_p$ for each point, we can calculate the pairwise similarities using the formula in equation \ref{Eq-003}, as shown in statements 10 to 14 in Algorithm \ref{Alg:Alg-01}. Large values indicate highly similar points, whereas low values indicate dissimilarity. We build another histogram of all the pairwise similarities using the Freedman–Diaconis rule \cite{RN387} as shown in Figure\ \ref{Fig:Fig-05}. For each point, similarities lower than the threshold $T_p$ are eliminated. If the maximum similarity is larger than the mean plus the standard deviation $\mu + \sigma$, the threshold is set as $T=\mu + \sigma$. If not, the threshold is set as $T=\mu - \sigma$. Figure\ \ref{Fig:Fig-05} shows the included similarities as blue bins and the excluded similarities as red bins. The graph edges are defined as:
\begin{equation}
(p,q) \in E(G) \Leftrightarrow A_{pq} > T_p.
\label{Eq-005}
\end{equation}
\noindent
where $(p,q)$ is the edge between points $p$ and $q$. $A_{pq}$ is the weight assigned to the edge $(p,q)$. This process is described in statements 15 to 21 in Algorithm \ref{Alg:Alg-01}.

The last step of our reduction method is to build a mutual graph. In a mutual graph, a pair of points should agree to accept an edge. This makes the graph $G$ to be defined as:
\begin{equation}
(p,q) \in E(G) \Leftrightarrow A_{pq} > T_p \quad \text{and} \quad A_{qp} > T_q.
\label{Eq-006}
\end{equation}
\noindent
where $T_p$ is threshold of acceptance for the vertex $p$.

\begin{figure}[!t]
	\removelatexerror
	\begin{algorithm}[H]
		\DontPrintSemicolon
		
		\KwInput{$k$-nn graph where $k=k_{max}$ of $N$ vertices.}
		\KwOutput{Reduced graph of $N$ vertices.}
		
		Construct distance matrix $D(N,k_{max})$ of $k$-nn graph
		
		Construct a histogram $H_D$ of all elements in $D$ using FD rule
		
		Save bin width in $H_D$ to the variable $bin_D$
		
		\tcc{The following loop has computations in order of $\mathcal{O}(Nk_{max})$}
		\For{$p = 1$ to $N$}
		{
			Construct a histogram $H_p$ of $D_{p,1 \text{ to } k_{max}}$ using $bin_D$
			
			Apply MWA to bin values in $H_p$ (equation \ref{Eq-004})
			
			Set K$^{\text{th}}$ as the first bin that exceeds MWA threshold
			
			$\sigma_p = \text{mean}(D_{p,1 \text{ to } \text{K}^{\text{th}}})$
			
		}
		
		\tcc{The following loop has computations in order of $\mathcal{O}(Nk_{max})$}
		\For{$p = 1$ to $N$}
		{
			\For{$q = 1$ to $k_{max}$}
			{
				$A_{p,q}=\exp{\left(\frac{D\left(p,q\right)}{\sigma_p \sigma_q\ }\right)}$
			}		
		}
		
		\tcc{The following loop has computations in order of $\mathcal{O}(Nk_{max})$}
		\For{$p = 1$ to $N$}
		{
			\uIf{$\text{max}(A_{p,1 \text{ to } k_{max}}) > \mu(A_{p,1 \text{ to } k_{max}})+\sigma(A_{p,1 \text{ to } k_{max}})$}
			{
				$A_{p,1 \text{ to } k_{max}} < \mu(A_{p,1 \text{ to } k_{max}})+\sigma(A_{p,1 \text{ to } k_{max}}) = 0$
			}
			\Else
			{
				$A_{p,1 \text{ to } k_{max}} < \mu(A_{p,1 \text{ to } k_{max}})-\sigma(A_{p,1 \text{ to } k_{max}}) = 0$
			}
		}
		
		Construct a reduced graph using affinity matrix $A(N,k_{max})$
		
		\caption{Reducing a $k$-nearest neighbor graph}
		\label{Alg:Alg-01}
	\end{algorithm}
\end{figure}

\subsection{Integration with SpectralNet}
\label{SpectralNetApproach}

Our graph filtering method can be seamlessly integrated to the newly proposed spectral clustering using deep neural networks (SpectralNet) \cite{RN360}. SpectralNet uses Siamese nets \cite{RN426} to learn affinities between data points. Siamese nets expect the user to label which pairs are positive and which are negative. An unsupervised pairing uses the $k$-nearest neighbors, where the nearest neighbors are positive pairs and the farthest neighbors are negative pairs. Our graph filtering can be used to obtain positive and negative pairs. It offers the advantage of setting the number of pairs per point dynamically. This cannot be achieved using $k$-nearest neighbors, where all the points are restricted to have a fixed number of positive pairs. Also, we do not have to set $k$ manually. We let our method assigns the positive and negative pairs for each point. A pseudocode illustrating the steps of the proposed method is shown in Algorithm \ref{Alg:Alg-01}.% Our algorithm consists of four parts. First, the distance matrix $D$ is prepared, and the bin width $H_D$ is computed for each point. Secondly, the algorithm keeps only the closest neighbors as shown in Fig.\ \ref{Fig:Fig-04}. In the third part the affinity matrix $A$ is computed. Finally, the final loop filters out the edges with low similarities as shown in Fig.\ \ref{Fig:Fig-05}. 
\section{Experiments and discussions}
\label{Experiments}

\begin{figure*}%[t]%[!htp]
	\centering
	\includegraphics[width=0.7\textwidth,height=20cm,keepaspectratio]{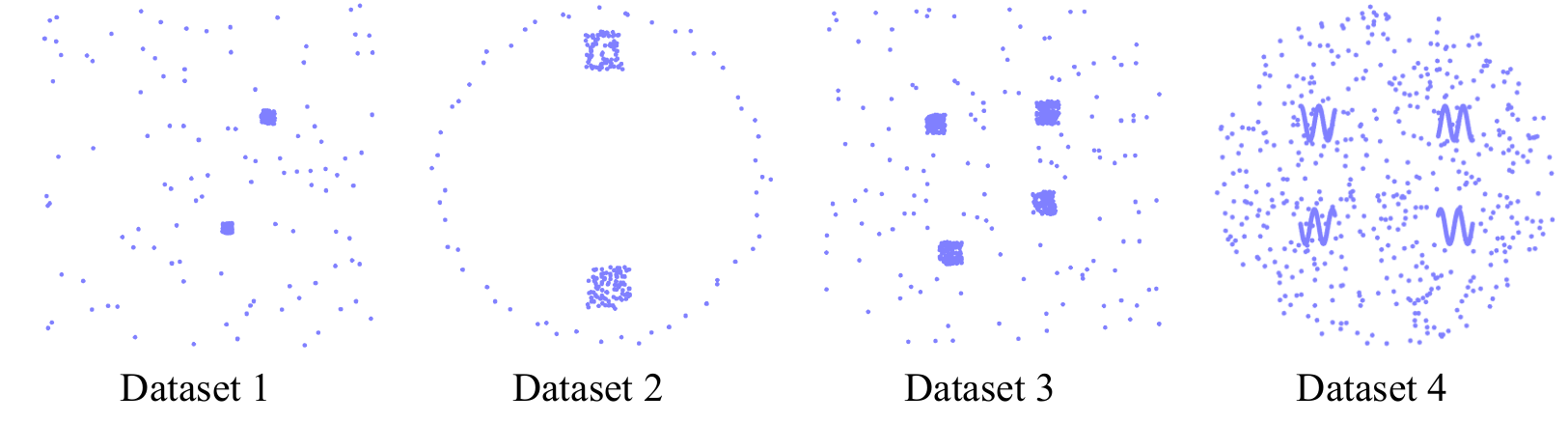}	
	\caption{Synthetic datasets used in the experiments.}
	\label{Fig:Fig-06}
\end{figure*}

In the experiments we used four synthetic datasets, as shown in Figure\ \ref{Fig:Fig-06}. \texttt{Dataset 1} to \texttt{3} were created by \cite{RN237}, while \texttt{Dataset 4} was created by us. We also used seven real datasets (see Table \ref{Table:Table-01}). Apart from the \texttt{MNIST} dataset, all the real datasets were retrieved from UCI machine learning. Each dataset was run with two parameter sets to evaluate the effect.% methods performed under different settings of parameters. The dimensionality of the \texttt{MNIST} dataset was reduced by PCA to 100 dimensions.
% UCI machine learning repository\footnote{\url{https://archive.ics.uci.edu/ml/index.php}}

\begin{table}%[h]%[!htp]
	\centering
	\caption{The four synthetic and seven real datasets used in the experiments; $N$ is the number of points, $d$ is the number of dimensions, $C$ is the number of clusters, $m$ is the size of the reduced set of vertices, and $\mu_0$ is the number of neighbors used as a threshold to include or exclude further neighbors.}
	\includegraphics[width=0.45\textwidth,height=20cm,keepaspectratio]{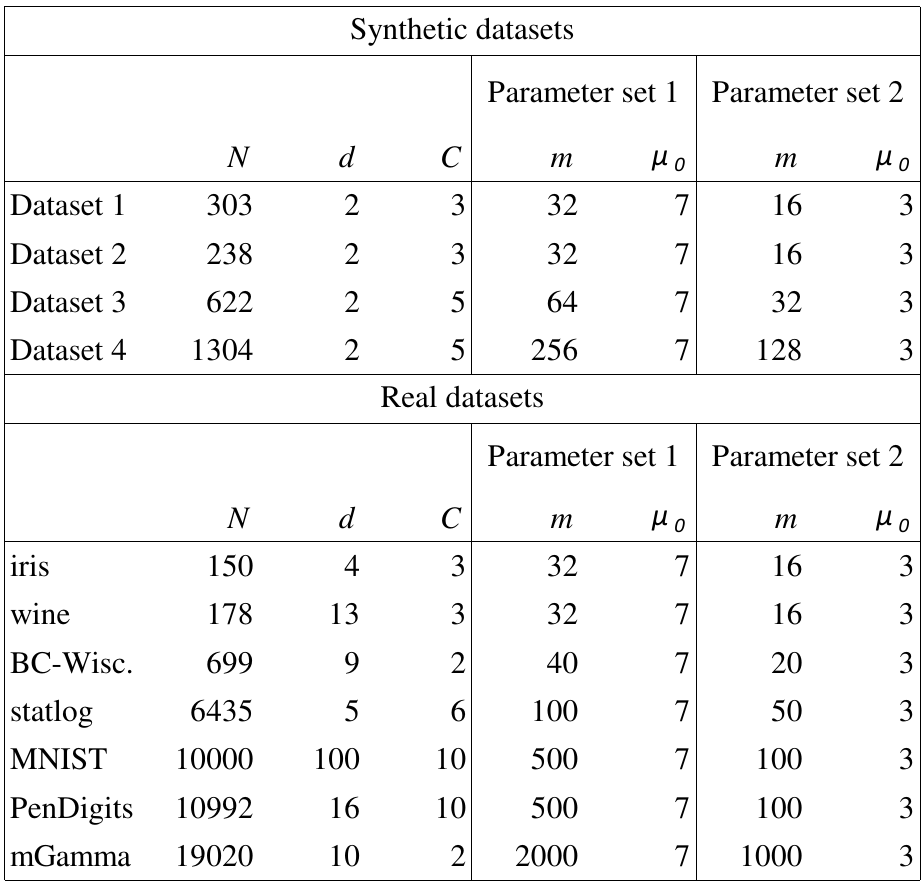}	
	\label{Table:Table-01}
\end{table}

Six methods were used for comparison and these are shown in Table \ref{Table:Table-02}. \texttt{Methods 1} to \texttt{5} \cite{RN232,RN275,RN276} rely on the parameter $m$, which is the number of representatives to build the graph $G$. They used iterative algorithms like $k$-means and self-organizing maps to construct the graph, which makes them produce a slightly different graph with each run. \texttt{Method 6} \cite{RN421} relied on the parameter $\mu_0$ to build the graph $G$, where $\mu_0$ is the number of neighbors whose mean was used as a threshold to include or exclude further neighbors. The code is available at \url{https://github.com/mashaan14/Spectral-Clustering}.

\begin{table*}%[h]%[!htp]
	\centering
	\caption{Methods used in the experiments. $m$ is the number of reduced vertices, $N$ is the number of all vertices, $t$ is the number of iterations, $k_{max}$ is the parameter used to construct $k$-nn graph.}
	\includegraphics[width=0.65\textwidth,height=20cm,keepaspectratio]{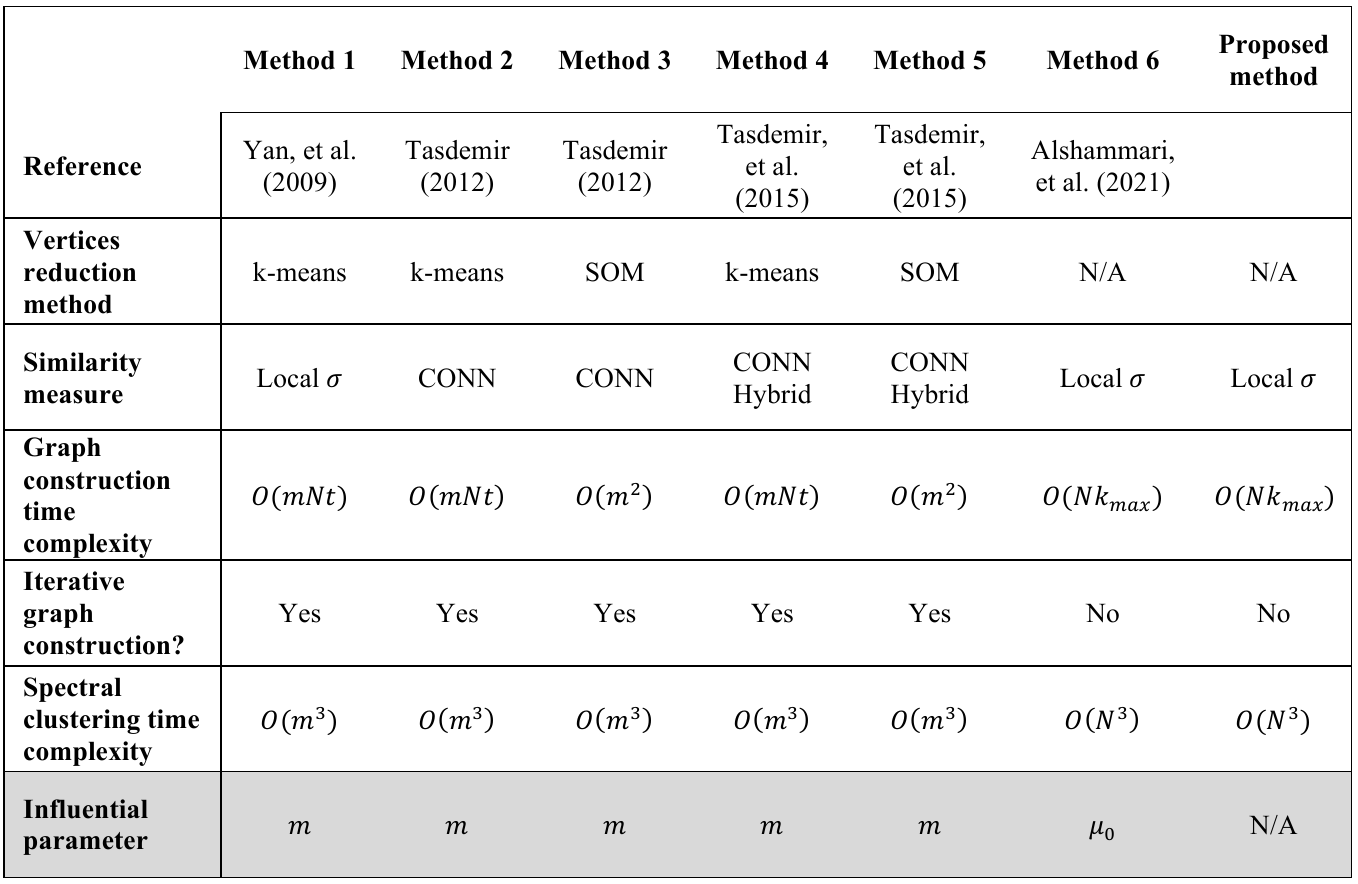}	
	\label{Table:Table-02}
\end{table*}

\begin{figure*}%[h]%[t]%[!htp]
	\centering
	\includegraphics[width=0.98\textwidth]{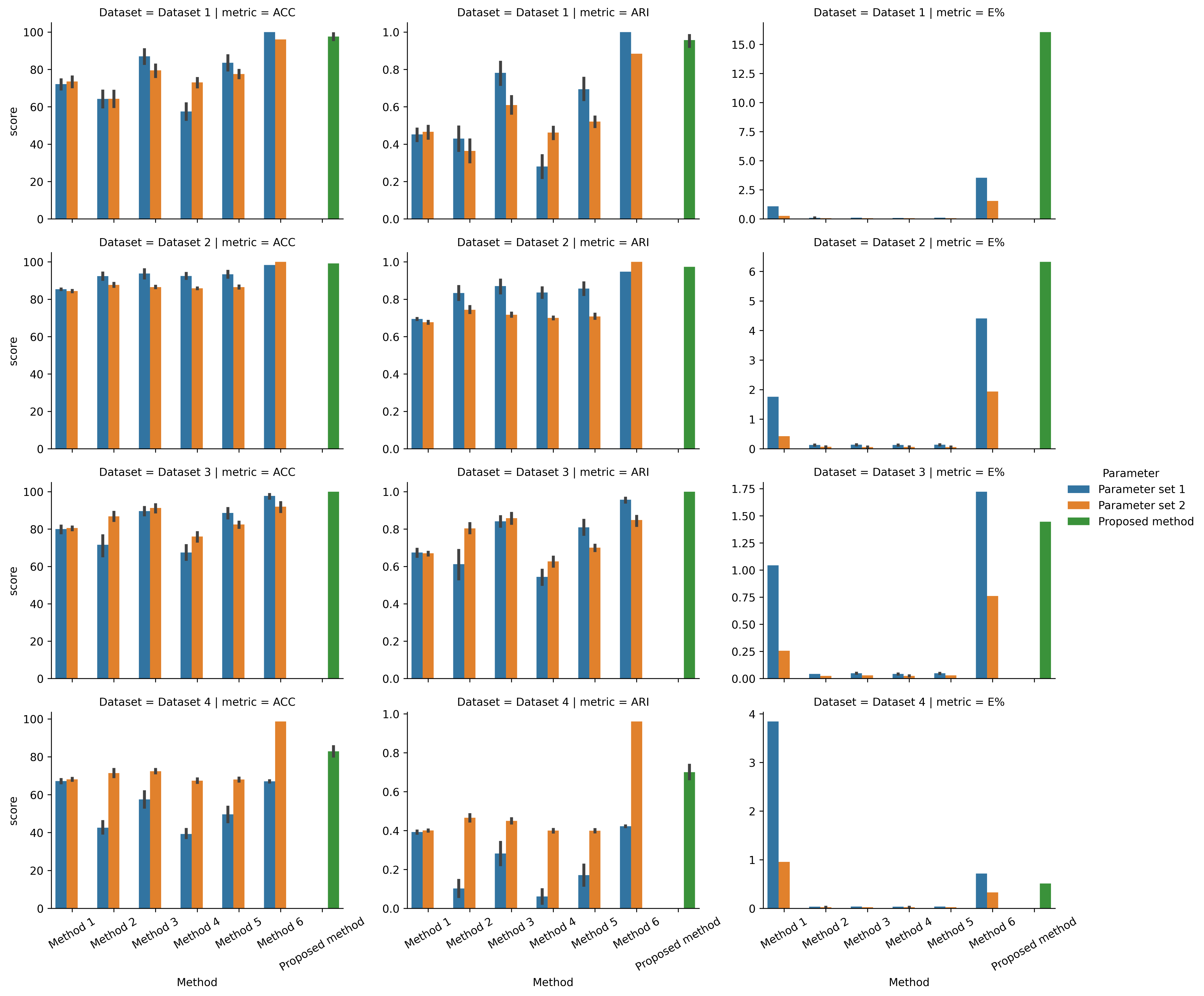} 
	\caption{Results with the synthetic data, all values are for 50 runs. (Best viewed in color)}	
	\label{Table:Table-03}
\end{figure*}

All the methods were evaluated using three evaluation metrics: 1) clustering accuracy (\textbf{ACC}) 2) the Adjusted Rand Index (\textbf{ARI}) \cite{RN365}, and 3) the percentage of edges used compared to all the edges in a full graph (\textbf{E\%}). 

ACC computes the percentage of hits between ground truth labels $T_i$ and labels obtained through clustering $L_i$. It is defined as \cite{Cai2005Document}:
\begin{equation}
	ACC(T,L)=\frac{\sum_{i=1}^{N}{\ \delta(T_i,map(L_i))}}{N},
	\label{Eq-ACC}
\end{equation}
\noindent
where $N$ is the number of points and the function $\delta(x,y)$ is the Kronecker delta function, which equals one if $x=y$ and zero otherwise. The function $map(L_i)$ permutes the grouping obtained through clustering for the best fit with the ground-truth grouping. ARI needs two groupings $T$ and $L$, where $T$ is the ground truth and $L$ is the grouping predicted by the clustering method. If $T$ and $L$ are identical, ARI produces one, and zero in case of random grouping. ARI is calculated using: $n_{11}$: pairs in the same cluster in both $T$ and $L$; $n_{00}$: pairs in different clusters in both $T$ and $L$; $n_{01}$: pairs in the same cluster in $T$ but in different clusters in $L$; $n_{10}$: pairs in different clusters in $T$ but in the same cluster in $L$.
\begin{equation}
ARI(T,L)=\frac{2(n_{00}n_{11}-n_{01}n_{10})}{(n_{00}+n_{01})(n_{01}+n_{11})+(n_{00}+n_{10})(n_{10}+n_{11})}\ .
\label{Eq-007}
\end{equation}
\noindent
The computational efficiency can be measured by the method’s running time, but, this is influenced by the type of machine used. We chose to measure the computational efficiency by the percentage of edges, \textbf{E\%}:
\begin{equation}
E\%=\frac{E(G_{reduced})}{E(G_{full})}\ .
\label{Eq-009-01}
\end{equation}

\subsection{Experiments on synthetic data}
\label{SyntheticData}
In the synthetic datasets the proposed method delivered a performance that ranked it as 2nd, 2nd, 1st, and 2nd for \texttt{Datasets 1} to \texttt{4} respectively (see Figure\ \ref{Table:Table-03}). \texttt{Method 6} was the top performer on three occasions. However, its performance dropped significantly when we changed the parameter $\mu_0$. For example, its performance dropped by 50\% with \texttt{Dataset 4} when we changed $\mu_0=3$ to $\mu_0=7$. This shows how parameters could affect the performance. Another observation is the consistency of ACC and ARI metrics over the 50 runs. By looking at Figure\ \ref{Table:Table-03}, the methods \texttt{1} to \texttt{5} have a wide standard deviation. This is explained by the iterative algorithms used by methods \texttt{1} to \texttt{5} to construct the graph. \texttt{Method 6} and the proposed method do not have this problem, and they have a small standard deviation. This is due to their deterministic nature when constructing the graph, which makes them consistent over independent executions.

In terms of the used edges, the proposed method used 6.32\%, 1.45\%, and 0.51\% of the full graph edges for \texttt{Datasets 2} to \texttt{4} respectively. But in \texttt{Dataset 1} there was a sharp increase where the proposed method used 16\% of the full graph edges. This sharp increase could be explained by the points in dense clusters being fully connected.

\begin{figure*}%[h]%[t]%[!htp]
	\centering
	\includegraphics[width=0.9\textwidth]{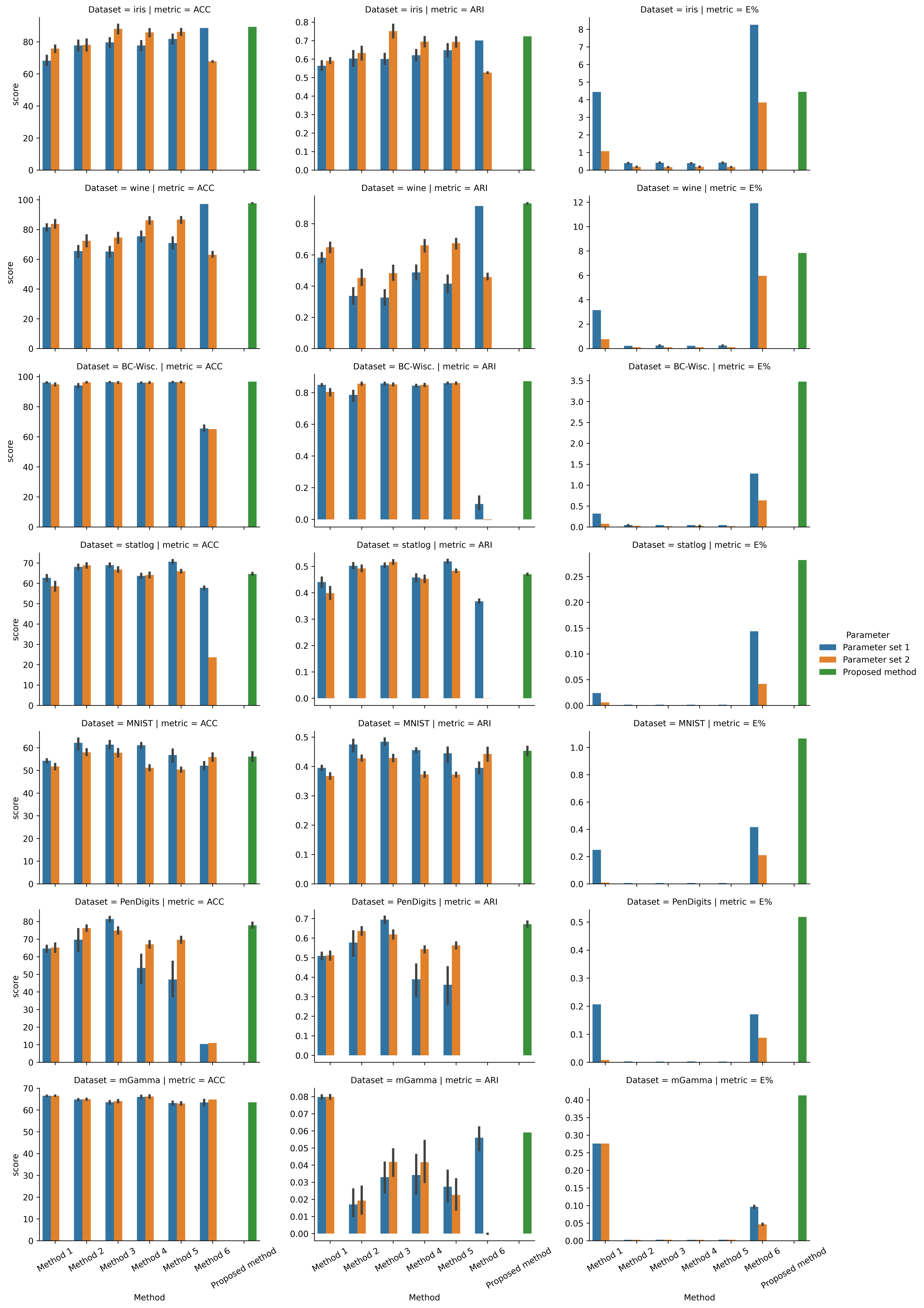} 
	\caption{Results with real data, all values are for 50 runs. (Best viewed in color)}	
	\label{Table:Table-04}
\end{figure*}

\subsection{Experiments on real data}
\label{RealData}
With real datasets in Figure\ \ref{Table:Table-04}, the proposed method continued to be the most consistent method over all tested methods. It kept a very small standard deviation, while other methods had a wide standard deviation. The performance of other methods was determined by their parameters. For example, \texttt{Method 3} was the best performer on \texttt{iris} dataset when $m=16$. However, when we changed $m$ to $32$ , its performance dropped by more than 15\%. Another observation with \texttt{statlog} and \texttt{MNIST} the proposed method did not perform well. This indicated that a cluster in these datasets does not have the same statistics across its regions. Therefore, characterizing clusters using local $\sigma$ might not be a good choice. Instead, we should use CONN to discover clusters discontinuity, rather than tracking local statistics.

\begin{figure*}%[h]%[t]
	\centering
	\includegraphics[width=0.48\textwidth]{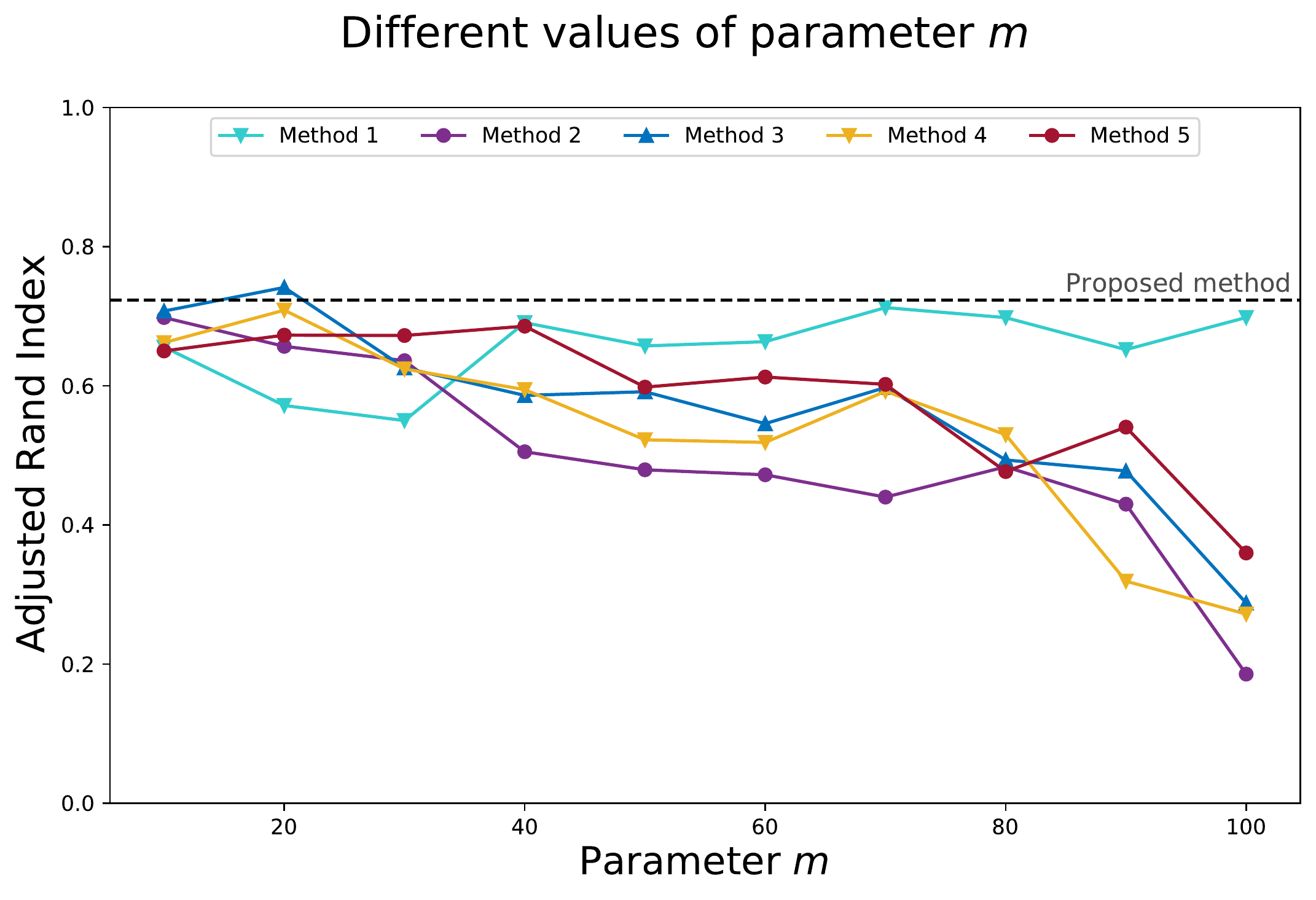}  
	\includegraphics[width=0.48\textwidth]{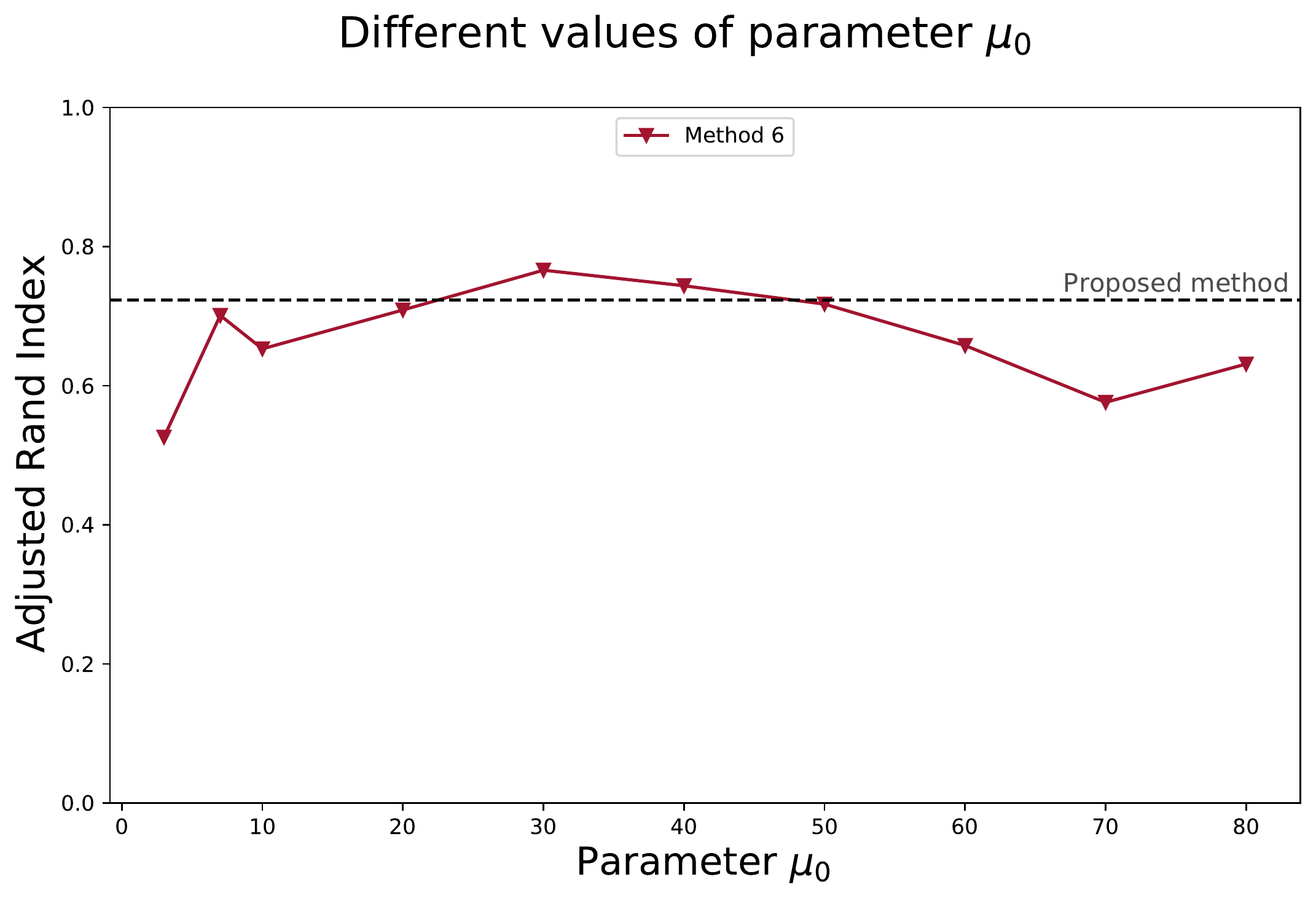}  
	\caption{Testing the methods’ performance with the \texttt{iris} dataset under different settings of parameters $m$ and $\mu_0$. (Best viewed in color)}
	\label{Fig:Fig-07}
\end{figure*}

\begin{figure*}%[h]%[!htp]
	\centering
	\includegraphics[width=0.7\textwidth,height=20cm,keepaspectratio]{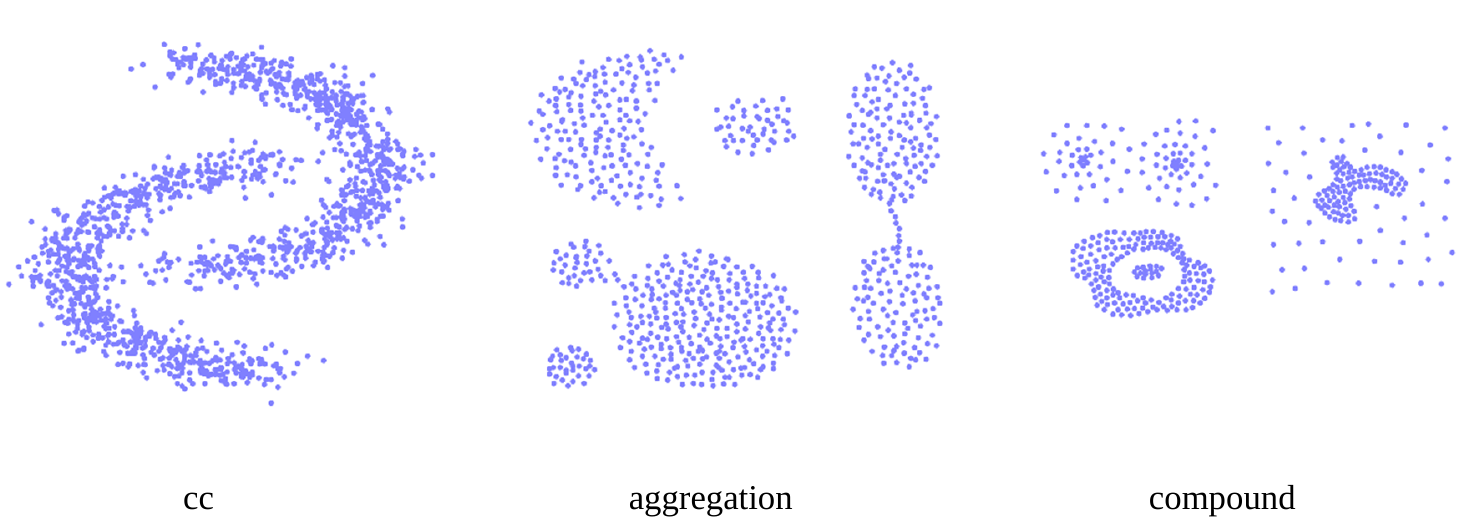}	
	\caption{Datasets used in the SpectralNet experiments.}
	\label{Fig:Fig-08}
\end{figure*}

\begin{figure*}%[h]%[!htp]
	\centering
	\includegraphics[width=0.75\textwidth,height=20cm,keepaspectratio]{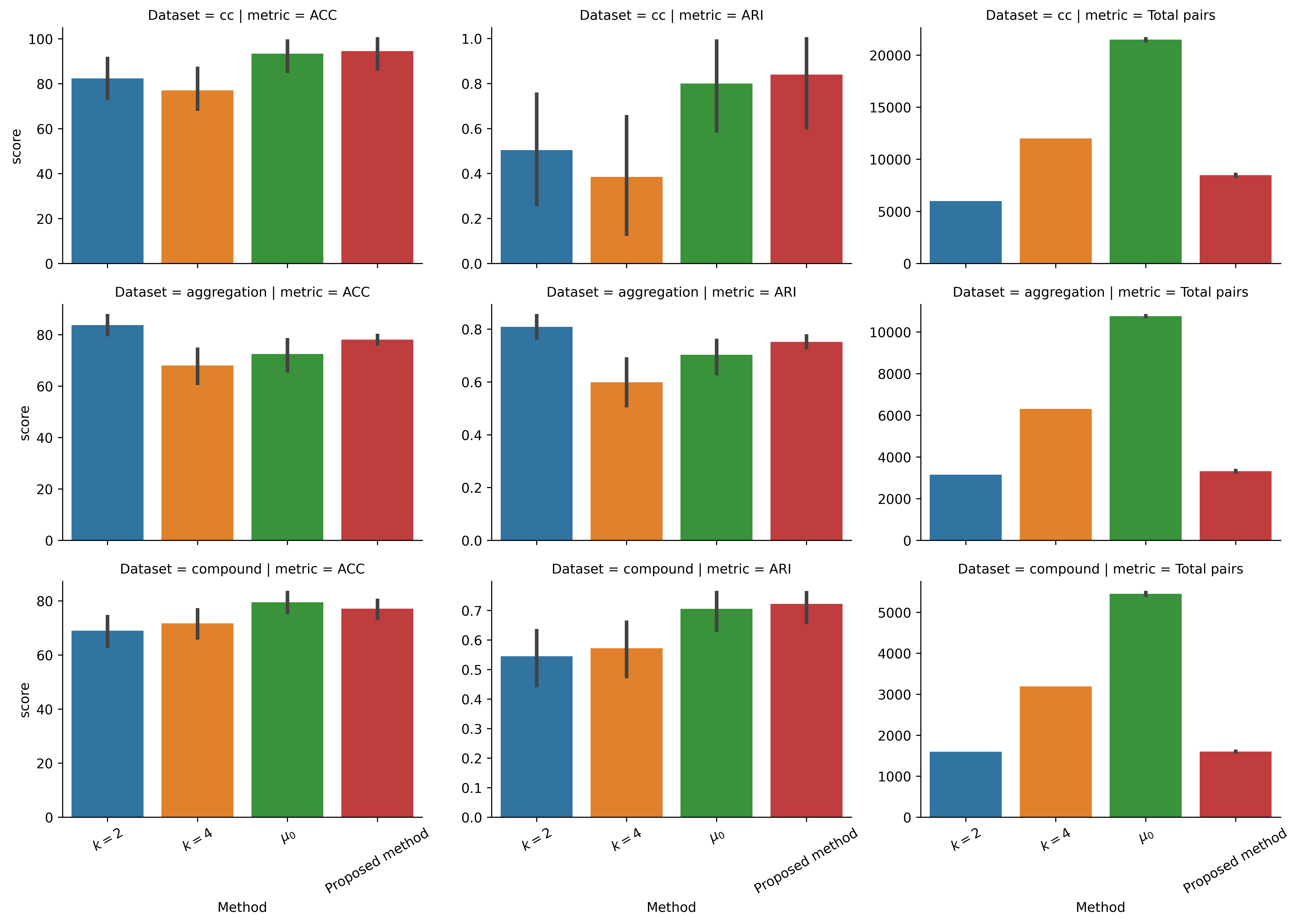}	
	\caption{Results of the experiments for integration with SpectralNet for 10 runs. (Best viewed in color)}	
	\label{Table:Table-05}
\end{figure*}

\subsection{Effect of the parameters on the spectral clustering performance}
\label{ParametersEffect}

In this experiment, we investigated how a wide selection of parameters could affect the accuracy of the spectral clustering. The parameters $m$ and $\mu_0$ were given the following values: $m \in \{ 10,20,30,40,50,60,70,80,90,100\}$ and $\mu_0 \in \{ 3,7,10,20,30,40,50,60,70,80\}$. In Figure\ \ref{Fig:Fig-07} (left), the performance \texttt{Methods 1} to \texttt{5} fluctuated with different values of $m$, with a clear downward trend seen as $m$ increased. The dashed horizontal line is the performance of the proposed method. In Figure\ \ref{Fig:Fig-07} (right), \texttt{Method 6} started with low performance, peaking around $\mu_0=30$, and then it took a downward trend. By eliminating the use of $\mu_0$, our method delivered a stable performance as shown by the horizontal dashed line.

\subsection{Experiments for integration with SpectralNet}
\label{SpectralNetExperiments}

The SpectralNet integration experiment was conducted using three datasets shown in Figure\ \ref{Fig:Fig-08}. The evaluation metrics are \textbf{ACC} shown in equation \ref{Eq-ACC}, \textbf{ARI} shown in equation \ref{Eq-007}, and \textbf{total pairs}, which is the number of pairs passed to the Siamese net. We used four methods to construct positive and negative pairs. The first two methods used a $k$-nearest neighbor graph with $k=2$ and $k=4$. Simply, the nearest $k$ neighbors were set as the positive pairs, and $k$ random farthest neighbors were set as the negative pairs. The third method used the parameter $\mu_0$ proposed by Alshammari, et al. \cite{RN421} to construct pairs. 

In Figure\ \ref{Table:Table-05}, the proposed method delivered the best performance for the \texttt{cc} and \texttt{compound} datasets. This good performance was coupled with good computational efficiency, with an average of $8,468$ for the total pairs passed to the Siamese net. Only $k=2$ could deliver fewer total pairs, but with a massive loss in performance. For the \texttt{aggregation} dataset, $k=2$ delivered the best performance. This experiment highlighted the need for setting the number of positive pairs dynamically. The methods following this approach (the $\mu_0$ method and our method) were the best performers for two of the three datasets. 

%\begin{figure*}%[!htp]
%	\centering
%	\includegraphics[width=0.68\textwidth,height=20cm,keepaspectratio]{figs-01/Fig-09.pdf}	
%	\caption{Sample results from the experiments for integration with SpectralNet. (Best viewed in color).}
%	\label{Fig:Fig-09}
%\end{figure*}

%Figure\ \ref{Fig:Fig-09} shows sample results from the experiments for integration with SpectralNet.

%\subsection{Advantages and limitations}
%\label{AdvantagesAndLimitations}
%The proposed method offers a number of advantages over the surveyed methods. First, it does not use iterative methods to construct the graph needed for spectral clustering and SpectralNet. These iterative methods use random initialization that makes their outcome different over independent executions. The proposed method is deterministic which makes it consistent over independent executions. Second, the proposed method does not have manually-tuned parameters to construct the graph. This makes spectral clustering and SpectralNet more automated.
%
%There are some limitations that need to be considered before implementing the proposed method. First, its graph construction time is $\mathcal{O}(Nk_{max})$, which is in terms of $N$. The time the surveyed methods take to construct the graph is in terms of $m$ where $m \ll N$. $m$ is the size of the reduced set of vertices. Another limitation of the proposed method is the reliance on the gaussian kernel to compute pairwise similarities. There are some recent studies that offer another way to compute pairwise similarities, like random projection forests \cite{8676336,yan2019similarity}.

\section{Conclusion}
\label{Conclusion}
The problem of detecting non-convex clusters has led to the development of numerous clustering methods. One of the well-known graph-based clustering methods is spectral clustering and SpctralNet. Both spectral clustering and SpectralNet require a graph that connects points in the same cluster with edges of high weights. The intuition is simple, strongly connected points will become closer in the embedding space and can be easily detected. 

Graph reduction requires extensive use of parameters that need careful setting for each dataset. The graph reduction algorithm proposed in this study does not require any parameters to reduce the graph, yet it is able to maintain spectral clustering and SpectralNet accuracies. It takes an input as a full graph or a $k$-nearest neighbor graph (in the case of a large number of points). Then, it reduces the graph edges using statistical measures that require low computations. The experiments revealed that the proposed method provides a stable alternative compared to other methods that require parameters tuning.

The proposed method does not reduce the graph vertices, which could boost the computational efficiency. A useful extension of the proposed method would be a vertices reduction component that is aware of local statistics. Another potential improvement of this work is to use a different kernel other than gaussian kernel to compute pairwise similarities.

\bibliographystyle{cas-model2-names}

\bibliography{mybibfile}

\bio{}
Dr. Mashaan holds a MSc in computer science from King Fahd University of Petroleum and Minerals (KFUPM), Saudi Arabia, and a PhD from the University of Sydney, Australia. His research interests lie broadly in graph clustering and deep representation learning.
\endbio

\bio{}
Dr. John Stavrakakis has strong interests in 3D computer graphics, remote rendering and computer security. He holds a PhD in Computer science and is an academic fellow at the University of Sydney, Australia.
\endbio

\bio{}
Dr. Masahiro Takatsuka received his MEng degree at Tokyo Institute of Technology in 1992, and received his PhD at the Monash University in 1997. In 1997-2002, he worked at GeoVISTA Center, The Pennsylvania State University as a senior research associate. He joined the School of Computer Science, the University of Sydney in 2002.
\endbio

\end{document}